%% file: paper.tex
\documentclass[]{bytedance_seed}

\usepackage[toc,page,header]{appendix}
\usepackage{amsfonts}
\usepackage{algorithm}
\usepackage{algpseudocode}

\usepackage{minitoc}

\title{FocusDPO: Dynamic Preference Optimization for Multi-Subject Personalized Image Generation via Adaptive Focus}

\author[*]{Qiaoqiao Jin}
\author[*\dagger]{Siming Fu}
\author[*]{Dong She} 
\author[]{Weinan Jia}
\author[]{Hualiang Wang}
\author[]{Mu Liu}
\author[\ddagger]{Jidong Jiang}

\affiliation[]{ByteDance FanQie}
\contribution[*]{Equal Contribution}
\contribution[\dagger]{Project Lead}
\contribution[\ddagger]{Corresponding Author}

\abstract{
Multi-subject personalized image generation aims to synthesize customized images containing multiple specified subjects without requiring test-time optimization. However, achieving fine-grained independent control over multiple subjects remains challenging due to difficulties in preserving subject fidelity and preventing cross-subject attribute leakage. We present \textbf{FocusDPO}, a framework that adaptively identifies focus regions based on dynamic semantic correspondence and supervision image complexity. During training, our method progressively adjusts these focal areas across noise timesteps, implementing a weighted strategy that rewards information-rich patches while penalizing regions with low prediction confidence. The framework dynamically adjusts focus allocation during the DPO process according to the semantic complexity of reference images and establishes robust correspondence mappings between generated and reference subjects. Extensive experiments demonstrate that our method substantially enhances the performance of existing pre-trained personalized generation models, achieving state-of-the-art results on both single-subject and multi-subject personalized image synthesis benchmarks. Our method effectively mitigates attribute leakage while preserving superior subject fidelity across diverse generation scenarios, advancing the frontier of controllable multi-subject image synthesis.}

\checkdata[Page]{\url{https://bytedance-fanqie-ai.github.io/FocusDPO/}}

\begin{document}
\maketitle
\input{sections/introduction}
\input{sections/relatedwork}

\input{sections/approach}
\input{sections/experiments}

\bibliographystyle{plainnat}
\bibliography{paper}

\clearpage

\beginappendix

\input{sections/appendix}

\end{document}

%% file: sections/introduction.tex
\section{Introduction}
The rapid advancement of diffusion models~\cite{ddpm} has revolutionized personalized image generation~\cite{dreambooth,textual-inversion,lora,ipadapter,blipdiffusion,uno,xverse,omnicontrol,chen2025customcontrast}, enabling the synthesis of high-quality images featuring specific subjects of interest. Among various personalization paradigms, multi-subject personalized image generation has emerged as a particularly compelling research direction, aiming to synthesize customized images containing multiple specified subjects without requiring computationally expensive test-time optimization. This capability holds significant practical value for applications ranging from creative content generation to personalized advertising and digital art creation.

The multi-subject personalization methods~\cite{uno, xverse, dreamo, omnicontrol, omnigen, kumari2023multi, realcustom, mao2024realcustom++} fundamental difficulty lies in achieving fine-grained independent control over multiple subjects while simultaneously preserving the visual fidelity of each individual subject. Existing approaches~\cite{uno, omnicontrol, omnigen} often struggle with cross-subject attribute confusion, where characteristics from one subject inadvertently influence the appearance of another, leading to inconsistent or corrupted generations. Moreover, maintaining the precise details and distinctive features of each reference subject becomes increasingly complex as the number of subjects grows, particularly when subjects share similar semantic categories or visual attributes.

\begin{figure*}[t]
\centering
\includegraphics[width=1.0\linewidth]{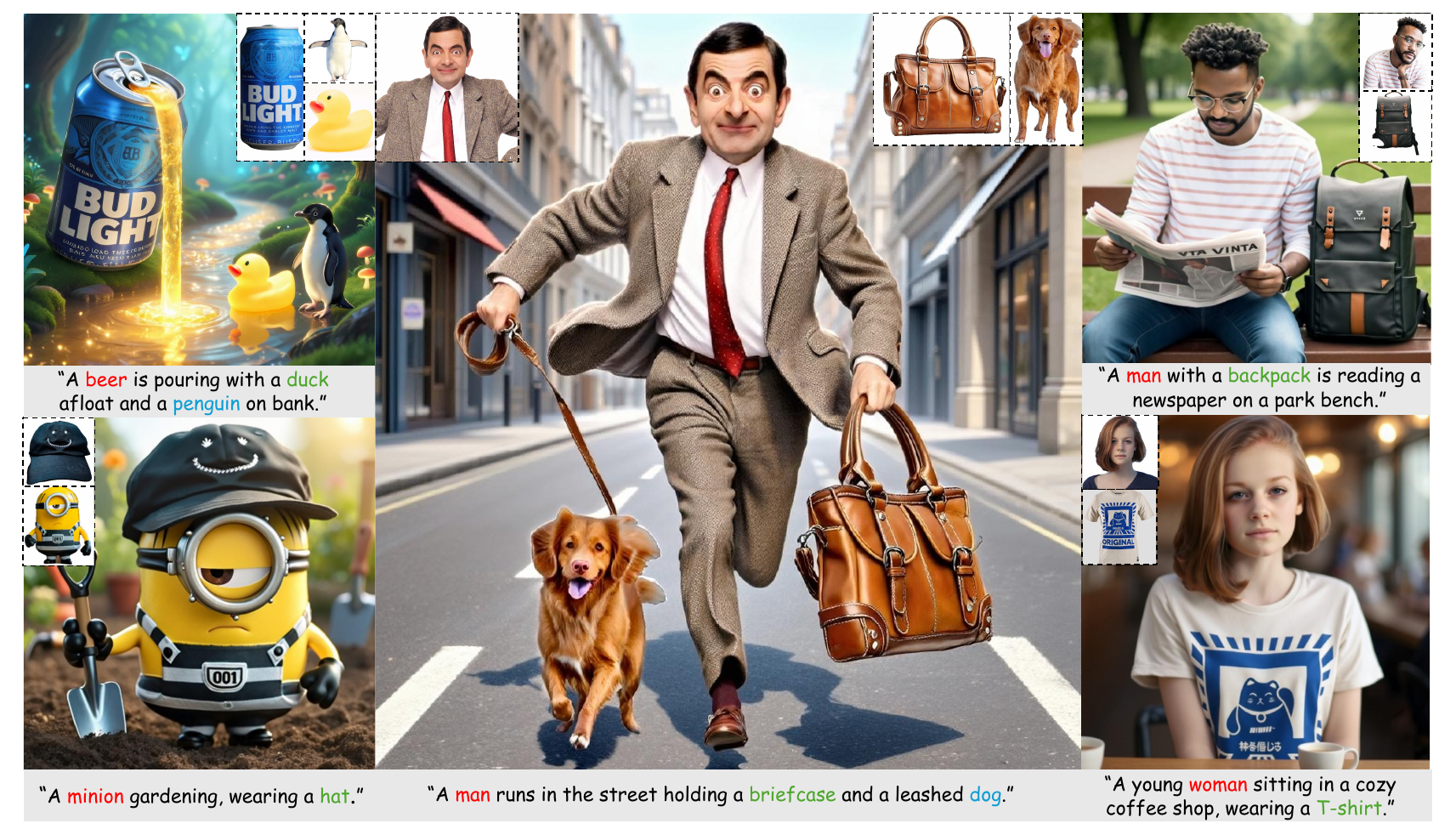}
\caption{\textbf{Our proposed FocusDPO demonstrates capabilities in single-subject and multi-subject driven generation tasks.}}
\label{fig:motivation}
\end{figure*}

\begin{figure*}[t]
\centering
\includegraphics[width=1.0\linewidth]{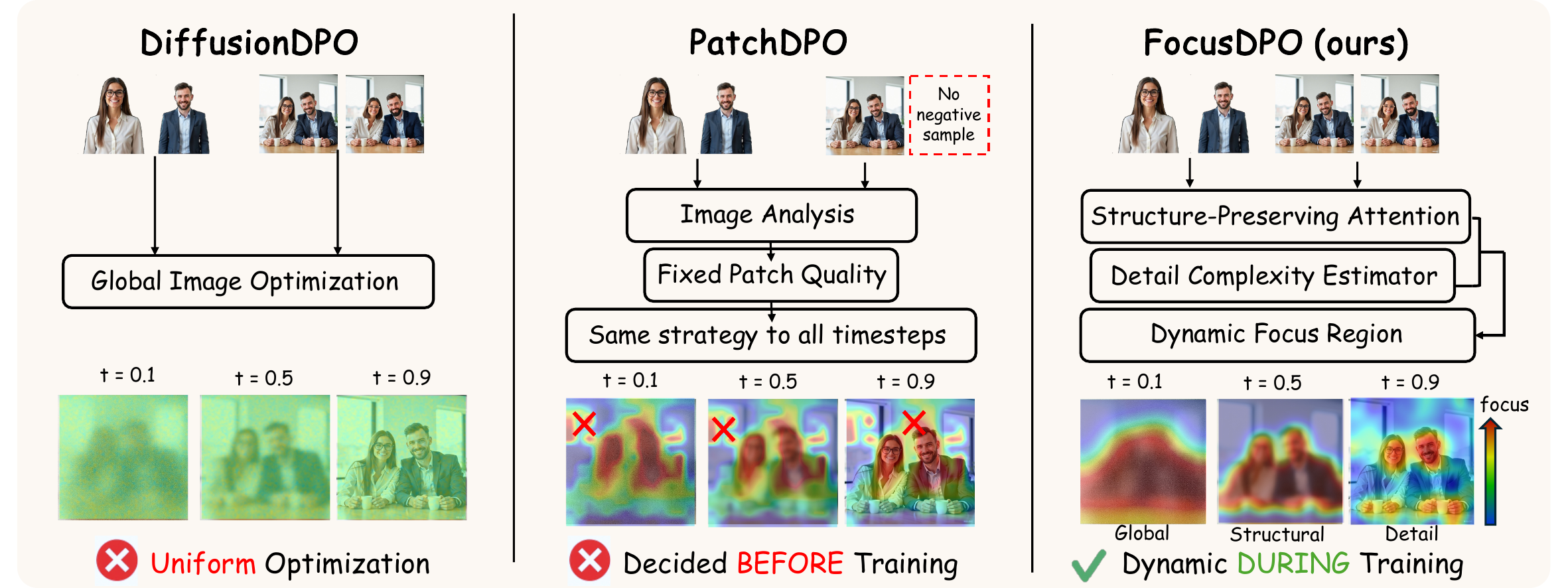}
\caption{\textbf{Comparison of training optimization strategy: DiffusionDPO's uniform optimization (left) vs. PatchDPO's fixed patch optimization (middle) vs.  FocusDPO's adaptive focus strategy (right).} }
\label{fig:motivation}
\end{figure*}

Recent efforts in multi-subject generation have explored various strategies, such as PatchDPO~\cite{patchdpo}, it estimates the quality of image patches within each generated image and accordingly trains the model. However, these methods~\cite{patchdpo, diffusiondpo} typically employ fixed treatment on different image regions across different training timesteps, failing to account for the varying complexity and semantic importance of different areas within the generated image. This limitation becomes particularly pronounced when dealing with subjects of different scales, positions, or semantic complexity levels. As shown in Fig.~\ref{fig:motivation}, when the noise strength is changed, the regions that require focused attention during model training should adapt accordingly. At higher noise levels, the model needs to concentrate on global structure and semantic features, while at lower noise levels, attention should shift toward fine-grained details and local texture preservation. \textbf{\textit{This observation motivates our dynamic focus modulation mechanism, which adjusts the spatial focus regions based on the current denoising step and the complexity of the supervision signal.}}

To address these challenges, we present FocusDPO, a novel framework that leverages dynamic Direct Preference Optimization to achieve superior multi-subject personalized image generation. Our \textbf{\textit{key insight}} is that effective multi-subject control requires adaptive attention allocation based on both the dynamic correspondence between generated and reference subjects and the semantic complexity of supervision images. Rather than applying uniform optimization pressure across all image regions, FocusDPO intelligently identifies regions of focus and adjusts the training dynamics accordingly. The core contribution of our approach lies in its weighted training strategy that rewards high-quality image patches while penalizing regions with low prediction confidence. This selective optimization enables the model to concentrate computational resources on challenging areas while maintaining efficiency in well-handled regions. Furthermore, our framework establishes robust correspondence mappings between generated and reference subjects, ensuring consistent identity preservation across diverse generation scenarios. Extensive experiments demonstrate that FocusDPO substantially enhances the performance of existing pre-trained personalized generation models, achieving state-of-the-art results on both single-subject and multi-subject personalized image synthesis benchmarks. The proposed framework effectively addresses attribute leakage while maintaining superior subject preservation, marking a significant step forward in controllable image generation. Our primary contributions are threefold:

\begin{itemize}
    \item  We introduce FocusDPO, which intelligently identifies ``focus regions" characterized by high semantic complexity and detailed-preserving generation difficulty. By adaptively intensifying optimization on these key areas, it efficiently enhances overall image quality and training stability.

    \item By dynamic semantic and detail-preserving preference optimization, our method effectively mitigates identity confusion and attribute leakage in multi-subject scenarios, ensuring faithful subject preservation.

    \item Extensive experiments demonstrate that FocusDPO substantially boosts the performance of existing  models, achieving state-of-the-art results on both single- and multi-subject personalized image synthesis benchmarks.
\end{itemize}

%% file: sections/relatedwork.tex
\section{Related Work}
\subsection{Subject-driven Generation}
Multi-subject personalized generation extends beyond single-subject methods like DreamBooth~\cite{dreambooth} and Textual Inversion~\cite{textual-inversion}, presenting significant challenges in maintaining subject integrity while preventing inter-subject interference. Recent approaches have explored various architectural strategies: OmniControl~\cite{omnicontrol} and IC LoRA ~\cite{iclora} everages DiTs as image encoders for subject references, while MS-Diffusion~\cite{msdiffusion} and MIP-Adapter~\cite{mipadapter} introduce specialized adapters for multiple subjects. UNO~\cite{uno} employs progressive training with Universal Rotary Position Embedding to mitigate attribute confusion, and XVerse~\cite{xverse} utilizes text-stream modulation for reference image processing. Despite these advances, existing methods struggle with inter-subject entanglement and attribute leakage, particularly when subjects share similar visual characteristics. This work addresses these limitations through dynamic semantic guidance, achieving robust multi-subject generation with enhanced semantic consistency.
\subsection{Diffusion-based Preference Optimization}
Direct Preference Optimization Direct Preference Optimization (DPO)~\cite{rafailov2023direct, zeng2024token, zhou2023beyond} and Reinforcement Learning from Human Feedback (RLHF)~\cite{bai2022training,casper2023open,fan2023dpok, lee2023aligning}, originally formulated for language model alignment, have been successfully adapted to diffusion-based image synthesis through methods including DPOK~\cite{fan2023dpok}, DDPO~\cite{black2023training}, DRaFT~\cite{clark2023directly} and AlignProp~\cite{prabhudesai2023aligning} for enhancing generation quality.
However, standard DPO frameworks fall short in consistency-sensitive tasks due to semantic confounds in global image comparisons. PatchDPO~\cite{patchdpo} mitigates this by optimizing preferences at the patch level, promoting finer-grained consistency learning. Unlike PatchDPO, which relies on static supervision without explicit positive-negative pairs, our method builds semantically aligned pairs and employs dynamic objectives to better preserve consistency.

%% file: sections/approach.tex
\section{Preliminary: Diffusion-DPO}

Our work builds upon Diffusion Direct Preference Optimization (Diffusion-DPO)~\cite{diffusiondpo}, a powerful framework for aligning text-to-image diffusion models with human preferences without requiring an explicit reward model. The training data for this process consists of preference pairs $\mathcal{D} = \{(x^w_0, x^l_0)\}$, where $x^w_0$ is the preferred (winning) image and $x^l_0$ is the dispreferred (losing) image for a given prompt. The core idea of Diffusion-DPO is to reinterpret the DPO loss in the context of diffusion model training:
\begin{equation}
\begin{split}
\mathcal{L}(\theta) &= -\mathbb{E}_{(x^w_0, x^l_0)\sim\mathcal{D},t\sim\mathcal{U}(0,T),x^w_t\sim q(x^w_t|x^w_0),x^l_t\sim q(x^l_t|x^l_0)} \\
&log\sigma(-\beta T\omega(\lambda_t)(||\epsilon^w-\epsilon_\theta(x^w_t,t)||^2_2 - ||\epsilon^w - \epsilon_\text{ref}(x^w_t,t)||^2_2 -(||\epsilon^l-\epsilon_\theta(x^l_t,t)||^2_2 - ||\epsilon^l - \epsilon_\text{ref}(x^l_t,t)||^2_2)),
\end{split}
\end{equation}
\noindent where $x^*_{t} = \alpha_t x^*_{0} + \sigma_t \epsilon^*$, $\epsilon^* \sim \mathcal{N}(0, I)$. And $\lambda_t = \frac{\alpha_t^2}{\sigma_t^2}$ is the signal-to-noise ratio. However, this uniform weighting strategy in Diffusion-DPO proves suboptimal for the nuanced demands of multi-subject generation. By treating all spatial regions with equal importance, the approach is inherently susceptible to interference from irrelevant background details. Consequently, its efficacy in preserving the consistency and identity of multiple subjects within a single composition is significantly constrained. To overcome this limitation, a more targeted weighting mechanism is necessary—one that can distinguish between foreground subjects and the background, focusing optimization where most needed.

\begin{figure*}[t]
\centering
\includegraphics[width=1.0\linewidth]{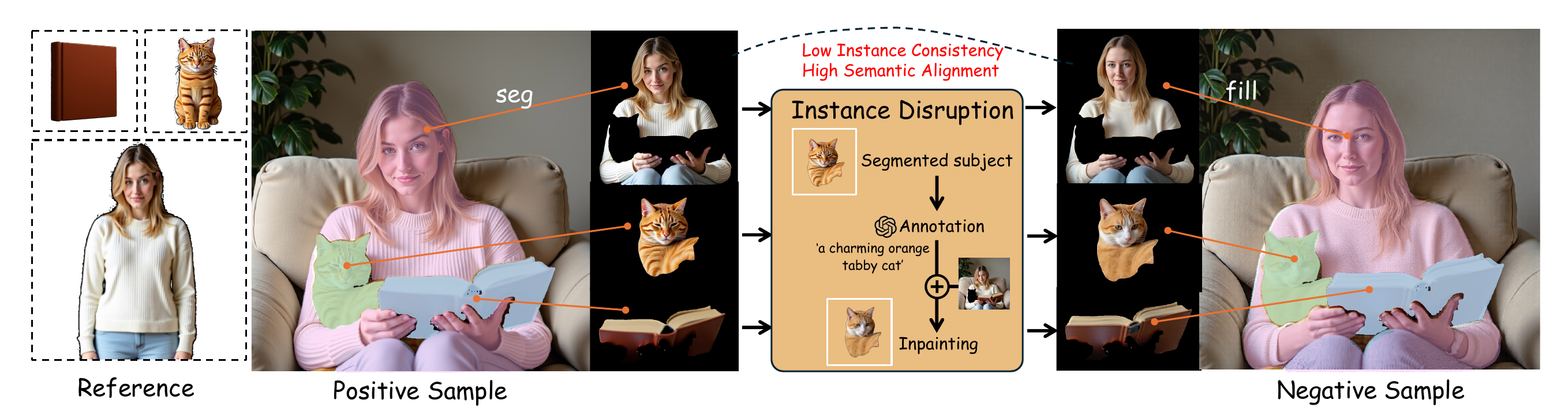} 
\caption{\textbf{Disrupted-Instance Pair Dataset (DIP) dataset construction workflow.} Preference pairs are generated by creating subject-consistent images, then introducing controlled perturbations to segmented areas to produce degraded counterparts.}
\label{fig:teaser}
\end{figure*}

\section{Method}
We propose \textbf{D}isrupted-\textbf{I}nstance \textbf{P}air Dataset (\textbf{DIP}) and Focus Direct Preference Optimization (FocusDPO), a spatially-aware preference optimization framework for addressing subject-level inconsistencies in personalized image generation. Our method constructs high-quality subject-consistent pairs $\{(x^w_0, x^l_0)\}$ with controlled subject variation, utilizing a binary prior guidance $\mathbf{M}_{\text{prior}} \in \{0, 1\}^{H\times W}$ to identify regions containing subject differences. We then introduce a spatial weighting mechanism $\mathbf{M} \in \mathbb{R}^{H\times W}$ that dynamically modulates the optimization process across different regions, where $H$ and $W$ denote the height and width of the latent feature maps.

\subsection{Disrupted-Instance Pair Dataset}
We construct the DIP Dataset as a semantically aligned collection of positive-negative image pairs designed to isolate subject-level inconsistencies. Each image pair maintains semantic content parity while introducing localized perturbations to subject regions. We begin by synthesizing reference and target pairs $\{(x_r, x_0^w)\}$ using the FLUX generator~\cite{flux}, conditioned on identical subject prompts $c$ but with varied auxiliary attributes such as background and lighting. From the resulting corpus, we manually curate 5,000 high-quality pairs for single-subject and multi-subject scenario, where $x_0^w$ shows strong visual alignment with $x_r$ in subject identity, serving as \textbf{positive samples}.

To construct \textbf{negative samples} $x_0^l$, we introduce controlled semantic disruptions to $x_0^w$ while preserving surrounding context. We apply GroundingSAM2~\cite{groundingdino, sam2} to extract accurate binary segmentation maps $\mathbf{M}_{\text{prior}}$ that isolate subject regions. To maintain semantic consistency in the altered regions, we utilize GPT-4o to generate fine-grained captions for the segmented areas, providing detailed semantic descriptions. These captions are fed into the inpainting model~\cite{flux}, which modifies subject-relevant pixels while preserving surrounding context. The resulting image $x_0^l$ exhibits degraded visual consistency with respect to $x_r$ while maintaining overall semantic coherence.

The final DIP dataset comprises quadruplets $(c, x_r, x_0^w, x_0^l)$, with shared conditioning prompt $c$ and reference image $x_r$ specifying target subject identity, a high-consistency image $x_0^w$ maintaining strong alignment with reference $x_r$, and a low-consistency counterpart $x_0^l$ generated through localized semantic perturbation within $\mathbf{M}_{\text{prior}}$. This approach ensures subject-level consistency in preference learning while removing confounders, forming the basis for spatially-aware optimization in FocusDPO.

\begin{figure*}[t]
\centering
\includegraphics[width=1.0\linewidth]{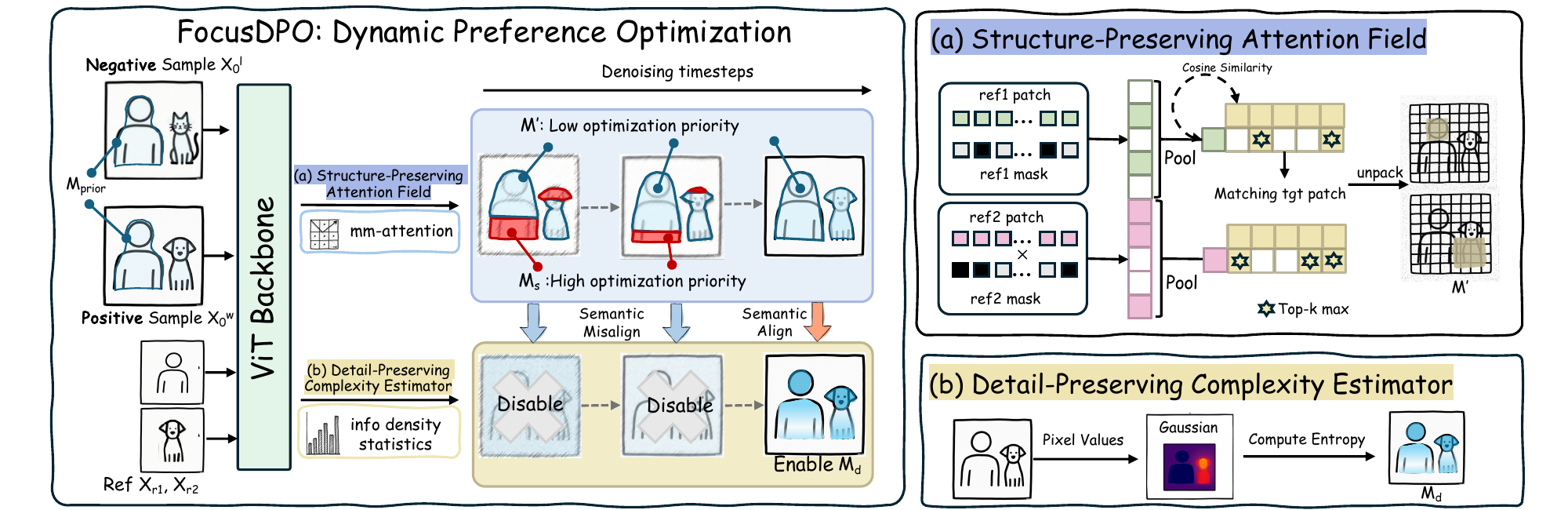} 
\caption{\textbf{The overall framework of FocusDPO.} FocusDPO introduce a \textit{spatially-aware optimization framework (left) }that adaptively focuses on critical regions through dynamic semantic guidance, leveraging (a) \textit{Structure-Preserving Attention Field} and (b) \textit{Detail-Preserving Complexity Estimator}.}
\label{fig:pipeline}
\end{figure*}

\subsection{Focus Direct Preference Optimization}
To overcome the limitations of uniform weighting in existing methods, we introduce a spatially-aware optimization framework, as shown in Fig.~\ref{fig:pipeline}. The cornerstone of our approach is a dynamic semantic guidance strategy that adaptively focuses supervisory signals on critical regions by decomposing the task into two complementary sub-problems: preserving global consistency via a Structure-Preserving Attention Field and maintaining high-fidelity features with a Detail-Preserving Complexity Estimator. We leverage these components to inform a spatial weighting mechanism that dynamically modulates the optimization process across semantically critical and non-critical regions.

\noindent \textbf{Structure-Preserving Attention Field:} Contemporary personalized generation methods suffer from subject confusion, where target attributes propagate to semantically unrelated regions, especially in multi-subject generation scenarios. Our primary objective is to address these structural deficiencies in generated images. Multi-modal attention layers inherently encode semantic relationships between the noised latent representation $x_t$ and the reference image $x_r$. For each attention layer $i$, we denote the target and reference token embeddings as:
\begin{equation}
\begin{cases}
H_{x_t}^{i} \in \mathbb{R}^{p_{x_t} \times d},  \\
H_{x_{r}}^{i} \in \mathbb{R}^{p_{x_\text{r}} \times d}, 
\end{cases}
\quad i \in \{0, 1, \ldots, N-1\},
\end{equation}
where $p_{x_t}$ and $p_{x_\textit{r}}$ denote the number of patch tokens, $d$ is the embedding dimension, and $N$ is the total number of multi-modal attention layers.

To establish semantic correspondences between subject regions and target patches, we employ a cross-layer correspondence strategy. For each reference subject patch $j$, we identify the target location with maximal semantic alignment by computing:
\begin{equation}
\begin{split}
&\mathcal{S} =  \sum_{i=1}^N \frac{1}{N}\cdot \langle  \frac{{CLS}_{x_\textit{r}}^{i}}{|{CLS}_{x_\textit{r}}^{i}|^2}, \frac{{H}_{x_t}^{i}}{|{H}_{x_t}^{i}|^2} \rangle, \\
& {CLS}_{x_\textit{r}}^{i} = \text{pool}({H}_{x_\textit{r}}^{i}), \quad {CLS}_{x_\textit{r}}^{i} \in \mathbb{R}^{1\times d}.
\end{split}
\label{eq3}
\end{equation}
Let $S$ be the score vector of size $1 \times p_{x_t}$, with elements $S_i$ for $i=1, \dots, p_{x_t}$. We identify the set of indices $\mathcal{J}$ corresponding to the $K$ largest scores in $S$:
\begin{equation}
    \mathcal{J} = \mathop{\arg\max}\limits_{i=1, \dots, p_{x_t}}(S_i),
\label{eq4}
\end{equation}
where $K$ is the number of tokens of reference image $x_r$ in $x_t$. Using this set, we construct a binary attention map $M'$ where element $\mathbf{M}'_j$ is set to 1 if its index $j$ is in the set of top-$K$ indices $\mathcal{J}$, and 0 otherwise:
\begin{equation}
    \mathbf{M}'_j = \mathbb{I}(j\in \mathcal{J}).
\label{eq5}
\end{equation}
Finally, we define Structure-Preserving Attention Field as:
\begin{equation}
    \mathbf{M}_s = \mathbf{M}_{\text{prior}} \setminus \mathbf{M}'.
\label{eq6}
\end{equation}

\begin{algorithm}[t]
\caption{FocusDPO Training Algorithm}
\label{alg:focusdpo}
\begin{algorithmic}[1] 

\Procedure{ComputeMasks}{$x_0^w, x_t^w, x_r$}
    \State Compute the Structure-Preserving Attention Field $\mathbf{M}_s$ based on Eq.~\ref{eq3}-Eq.~\ref{eq6}.
    \State Compute the Detail-Preserving Complexity Estimator $\mathbf{M}_d$ based on Eq.~\ref{eq7} and Eq.~\ref{eq8}.
    \State \textbf{return} $\mathbf{M}_s, \mathbf{M}_d$
\EndProcedure

\Statex 

\For{each training step}
    \State Sample preference pair $(x_0^w, x_0^l, \mathbf{M}_{\text{prior}}) \sim \mathcal{D}$, timestep $t \sim \mathcal{U}(1, T)$, noise $\epsilon \sim \mathcal{N}(0, I)$.
    \State Create noised latents $x_t^w, x_t^l$.

    \Statex 
    \Statex \textbf{Step 1: Generate dynamic semantic masks}
    \State $\mathbf{M}_s, \mathbf{M}_d \gets \Call{ComputeMasks}{x_0^w, x_t^w, x_r}$
    
    \Statex 
    \Statex \textbf{Step 2: Calculate the focus coverage ratio} 
    \State $A_{\text{focus}} \gets \|\mathbf{M}_s\|_1 / \|\mathbf{M}_{\text{prior}}\|_1$
    
    \Statex 
    \Statex \textbf{Step 3: Determine fusion mask} $\mathbf{M}$
    \State Determine fusion mask $\mathbf{M}$ using Eq.~\ref{eq9}.
    
    \Statex 
    \Statex \textbf{Step 4: Compute loss and update model}
    \State Compute the final loss $\mathcal{L}_{\text{FocusDPO}}$ using the spatially weighted objective (Eq.~\ref{eq10}).
    \State Update model parameters: $\theta \gets \theta - \eta \nabla_\theta \mathcal{L}_{\text{FocusDPO}}$.
\EndFor
\end{algorithmic}
\end{algorithm}

\noindent \textbf{Detail-Preserving Complexity Estimator:} Conventional preference optimization methods treat preference distinctions as globally distributed properties. However, for multi-subject generation in complex scenes, preference signals are often spatially localized. We observe that preference-critical regions correlate with areas of high visual complexity, which also pose challenges for diffusion model reconstruction. Motivated by this observation, we introduce a Detail-Preserving Complexity Estimator that prioritizes these complex regions during preference optimization through an information density weighting mechanism.

Our method generates this weight mask through two core steps. First, we compute a visual complexity score for each local region. For a patch centered at pixel $p$, denoted as $\text{patch}_p$, we obtain its complexity score $\mathbf{C}_p$ by calculating the Shannon entropy of its grayscale pixel intensity distribution~\cite{Jia_2025_CVPR}. Higher entropy values signify richer textures and details corresponding to higher visual complexity, while smooth and uniform regions yield lower entropy:
\begin{equation}
    \mathbf{C}_p = \text{CalculateComplexity}(\text{patch}_p). \label{eq7}
\end{equation}
Second, to ensure these scores are comparable and form a usable weighting scheme, we perform global normalization. We scale the complexity score $\mathbf{C}_p$ of each location to the range $[0, 1]$ to obtain the final information complexity $\mathbf{M}_d$:
\begin{equation}
    \mathbf{M}_d = \frac{\mathbf{C}_p - C_{\min}}{C_{\max} - C_{\min}}, 
\label{eq8}
\end{equation}
where $C_{\min}$ and $C_{\max}$ represent the minimum and maximum complexity scores computed across all local regions of the entire image. The resulting matrix of weights $\mathbf{M}_d$ constitutes our Detail-Preserving Complexity Estimator.

By incorporating this complexity estimator into the model's optimization loss function, we guide the model to focus attention on regions with higher weights during training. This enables the model to preferentially learn from and correct errors in critical areas such as fine textures and facial details, thereby significantly enhancing the local quality and overall fidelity of generated images.

\noindent \textbf{Final Loss:} As shown in Alg.~\ref{alg:focusdpo}, we determine the dynamic fusion field $\mathbf{M}$ through an adaptive strategy that leverages both structural and detail-preserving components. The fusion mechanism employs a focus threshold $\tau=0.1$ and tradeoff parameter $\gamma = 0.3$ to balance global consistency and local fidelity (\textit{with ablation study in appendix}):
\begin{equation}
    \mathbf{M} = \begin{cases}
        \mathbf{M}_s, & A_{\text{focus}} > \tau \\
         \gamma \mathbf{M}_s + (1 - \gamma)\mathbf{M}_d \odot \mathbf{M}_{\text{prior}}, & A_{\text{focus}} \leq \tau
        \end{cases}
\label{eq9}
\end{equation}
The complete FocusDPO objective incorporates the spatial fusion field into the preference optimization framework:
\begin{equation}
\begin{split}
\mathcal{L}_{\text{FocusDPO}}(\theta) &= -\mathbb{E}\log\sigma(-\beta T\omega(\lambda_t)(||(\epsilon^w-\epsilon_\theta(x^w_t,t))\odot\mathbf{M}||^2_2 - ||(\epsilon^w - \epsilon_\text{ref}(x^w_t,t))\odot\mathbf{M}||^2_2 \\
&-(||(\epsilon^l-\epsilon_\theta(x^l_t,t))\odot\mathbf{M}||^2_2 - ||(\epsilon^l - \epsilon_\text{ref}(x^l_t,t))\odot\mathbf{M}||^2_2)).
\end{split}
\label{eq10}
\end{equation}
This spatially-aware framework enables the model to preferentially learn from and correct errors in critical regions, significantly enhancing local quality and overall fidelity of generated images while maintaining subject consistency across complex multi-subject scenarios.

%% file: sections/experiments.tex
\begin{figure*}[t]
\centering
\includegraphics[width=1.0\linewidth]{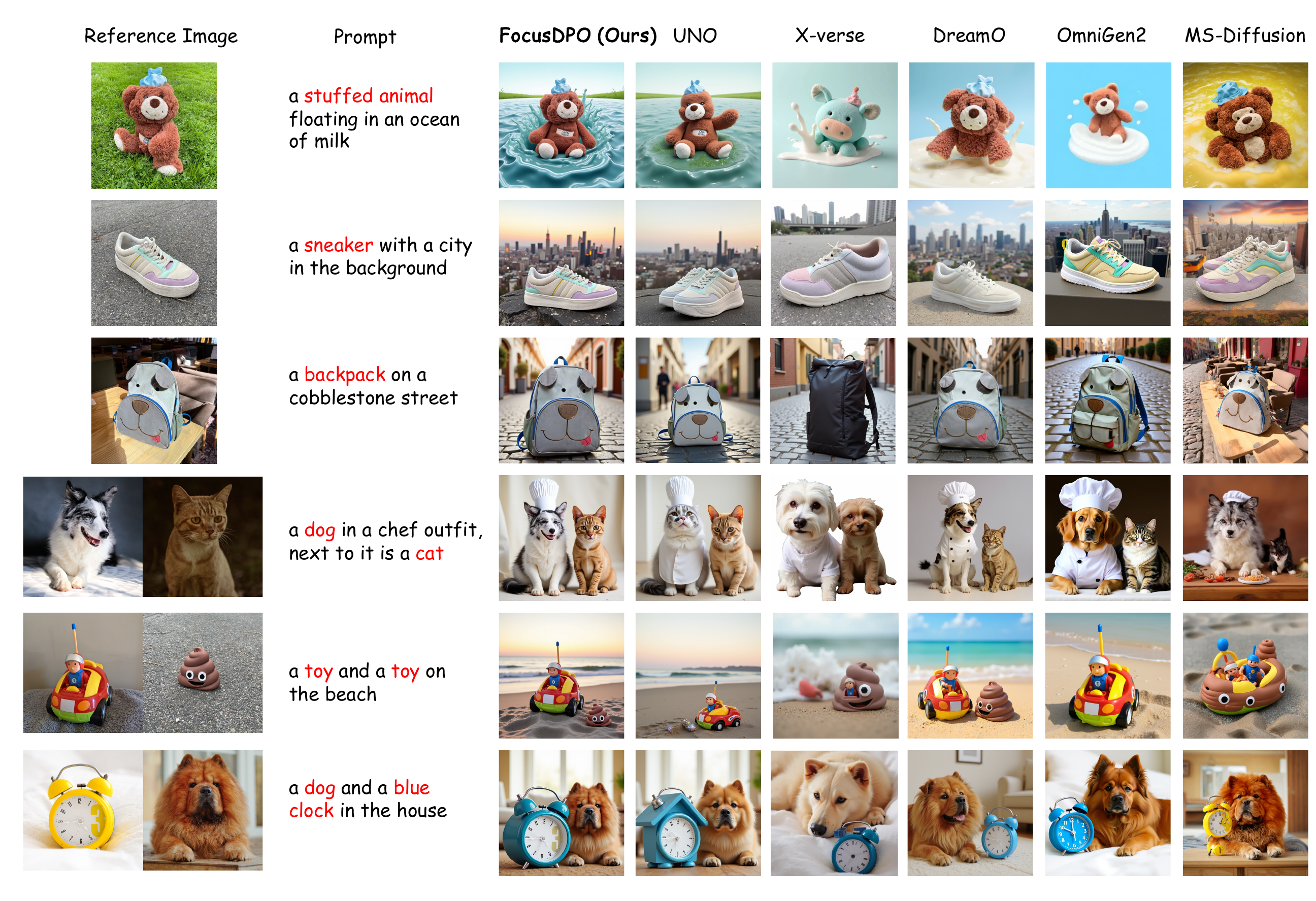}
\caption{Qualitative comparison of single-subject and multi-subject generation with different methods on DreamBench.}
\label{comp}
\end{figure*}

\section{Experiments}
\subsubsection{Implementation Details}
We implement our FocusDPO methodology on two representative diffusion architectures: U-Net~\cite{rombach2022high} and DiT~\cite{peebles2023scalable}, ensuring broad applicability and generalizability. For SDXL, we use the pre-trained IP-Adapter-Plus~\cite{ipadapter} with the SDXL~\cite{sdxl} text-to-image model, consistent with PatchDPO~\cite{patchdpo}. For DiT, we integrate FocusDPO into the UNO~\cite{uno} pipeline for multi-subject synthesis, utilizing FLUX.1 dev~\cite{flux} as the foundation model.

Our training uses rank-32 LoRA~\cite{lora} modules in both backbones, trained for 2000 steps on 8 GPUs with AdamW (lr=1e-8, batch size=1 per GPU). The training dataset DIP contains 10,000 preference pairs (5000 each for single- and multi-subject generation). The test dataset is DreamBench~\cite{dreambooth}, which includes both single-subject and multi-subject cases, with the multi-subject evaluation protocol following settings in~\cite{subjectdiffusion,mipadapter,uno}.

\subsubsection{Evaluation Metrics}
We evaluate subject fidelity and text-image alignment using established metrics. Subject fidelity is measured through CLIP-I~\cite{clip} and DINO~\cite{dino} scores, computing cosine similarity between generated, and reference images and text-image alignment is assessed using CLIP-T score.

\begin{table}[t]
\centering

\label{tab:clip_dino_compression}
\renewcommand{\arraystretch}{1.0}
\setlength{\tabcolsep}{5pt}
\begin{tabular}{l|ccc}
\toprule
\textbf{Method} & \textbf{DINO} $\uparrow$& \textbf{CLIP-I} $\uparrow$ & \textbf{CLIP-T} $\uparrow$  \\ \midrule
DreamBooth~\cite{dreambooth} & 0.668 & 0.803 & 0.305 \\
SSR-Encoder~\cite{ssrencoder} & 0.612 & 0.821 & 0.308 \\
RealCustom++~\cite{mao2024realcustom++} & 0.702 & 0.794 & \textbf{0.318} \\
OmniGen~\cite{omnigen} & 0.693 & 0.801 & 0.315 \\
OminiControl~\cite{ominicontrol} & 0.684 & 0.799 & 0.312 \\
DreamO~\cite{dreamo} & 0.712 & 0.809 & 0.314 \\
\midrule
IP-Adapter-Plus~\cite{ipadapter} & 0.692 & 0.826 & 0.281  \\
IP-Adapter-Plus + SFT & 0.691 & 0.828 & 0.279  \\
IP-Adapter-Plus + DPO & 0.695 & 0.831 & 0.276 \\
IP-Adapter-Plus + PatchDPO & 0.727 & 0.838 & 0.292 \\
IP-Adapter-Plus + FocusDPO & 0.751 & \underline{0.840} & 0.303\\
\midrule
UNO~\cite{uno} & 0.760 & 0.835 & 0.304  \\
UNO + SFT & 0.761 & 0.832 & 0.299   \\
UNO + DPO & \underline{0.764} & 0.835 & 0.301 \\
UNO + FocusDPO & \textbf{0.802} & \textbf{0.842} & \underline{0.316} \\
\bottomrule
\end{tabular}
\caption{Performance comparison for single-object personalized generation on DreamBench~\cite{dreambooth}. The best result is marked in \textbf{bold}, and the second-best is \underline{underlined}.}
\label{tab:1}
\end{table}

\begin{table}[t]
\centering
\label{tab:multi_subject_results}
\renewcommand{\arraystretch}{1.0}
\setlength{\tabcolsep}{10pt}
\begin{tabular}{l|ccc}
\toprule
\textbf{Method} & \textbf{DINO} $\uparrow$ & \textbf{CLIP-I} $\uparrow$ & \textbf{CLIP-T} $\uparrow$ \\
\midrule
DreamBooth~\cite{dreambooth} & 0.430 & 0.695 & 0.308 \\
MIP-Adapter~\cite{mipadapter} & 0.482 & 0.726 & 0.311 \\
MS-Diffusion~\cite{msdiffusion} & 0.525 & 0.726 & 0.319 \\
OmniGen~\cite{omnigen} & 0.511 & 0.722 & \textbf{0.331} \\
DreamO~\cite{dreamo} & 0.539 & 0.727 & 0.313 \\
\midrule
UNO (DiT Backbone) & 0.542 & \underline{0.733} & 0.322 \\
UNO + SFT & 0.542 & 0.732 &  0.321 \\
UNO + DPO & \underline{0.545} & 0.731 & 0.322 \\
UNO + FocusDPO & \textbf{0.570} & \textbf{0.739} & \underline{0.328} \\
\bottomrule
\end{tabular}
\caption{Multi-subject sythesis results on DreamBench. }
\label{tab:2}
\end{table}

\subsection{Qualitative Analysis}
We conduct a comprehensive comparative evaluation of our proposed methodology against existing state-of-the-art approaches across both single-subject and multi-subject generation paradigms. Our comparative analysis encompasses five recent and competitive methods: Xverse~\cite{xverse}, DreamO~\cite{dreamo}, OmniGen2~\cite{omnigen2}, MS-Diffusion~\cite{msdiffusion}, and our backbone architecture UNO~\cite{uno}. As demonstrated in Fig.~\ref{comp}, FocusDPO exhibits superior performance in maintaining subject consistency and identity preservation while simultaneously ensuring global semantic coherence.
The performance gains arise from our training paradigm, which dynamically targets complex, detail-rich regions to better capture subject-specific features and preserve fine-grained identity—especially in multi-subject generation.

\subsection{Quantitative comparisons} Tables~\ref{tab:1} and~\ref{tab:2} present comprehensive quantitative evaluations of our method against competing approaches on the DreamBench dataset for single-object and multi-subject personalized generation, respectively.We conduct evaluations of FocusDPO across two distinct backbone architectures, demonstrating consistent performance gains across all metrics. When integrated with the UNO backbone, FocusDPO establishes new state-of-the-art results, achieving a 5.5\% improvement on DINO (0.802 vs. 0.760) for single-object generation, and state-of-the-art performance (DINO: 0.570, CLIP-I: 0.739) for multi-subject synthesis. These empirical results substantiate the efficacy of our proposed framework.

\begin{figure}[t]
\centering
\includegraphics[width=0.87\linewidth]{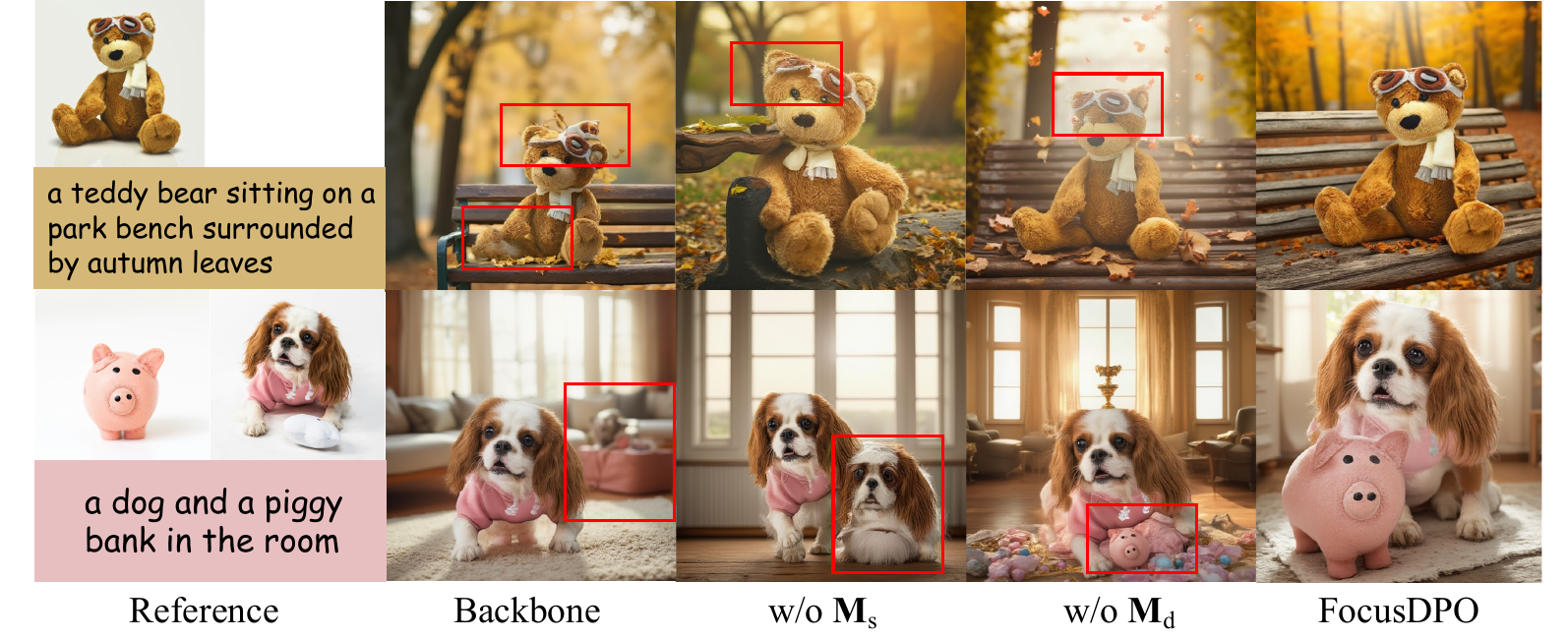} 
\caption{Ablation study on dynamic weighting components.}
\label{fig:ablation_components}
\end{figure}

\begin{figure}[t]
\centering
\includegraphics[width=0.88\linewidth]{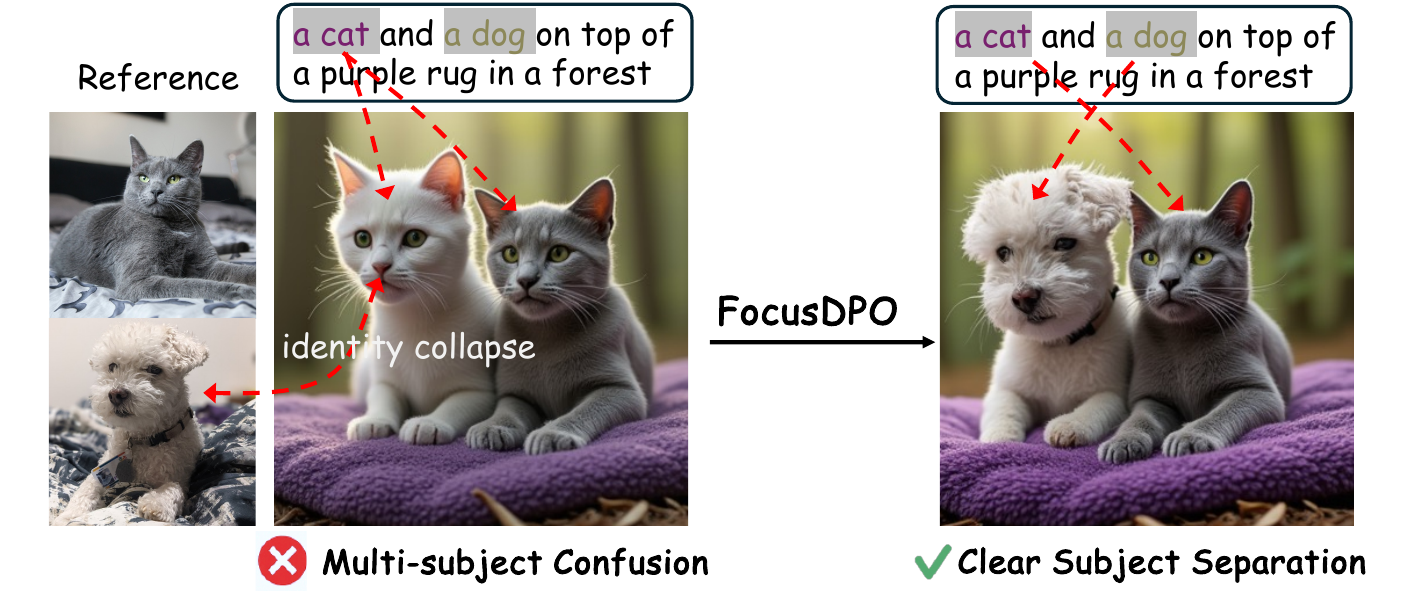} 
\caption{Analysis of resolving semantic confusion.}
\label{fig:confusion}
\end{figure}

\subsection{Ablation Study} 
\subsubsection{Ablation on Dynamic Fusion Components}
In Fig.~\ref{fig:ablation_components}, We conduct an ablation study to analyze the contribution of two critical weighting components within our dynamic optimization framework: the Structure-Preserving Attention Field \(\mathbf{M}_{s}\) and the Detail-Preserving Complexity Estimator \(\mathbf{M}_{d}\). The supplementary experimental results presented in Fig.~\ref{fig:confusion} demonstrate that the absence of \(\mathbf{M}_{s}\) results in substantial subject disambiguation failures, with this degradation being particularly pronounced in multi-subject synthesis scenarios. Conversely, the omission of \(\mathbf{M}_{d}\) manifests as inconsistencies in fine-grained structural coherence across the generated outputs.These results confirm that \(\mathbf{M}_{s}\) and \(\mathbf{M}_{d}\) work in tandem to preserve semantic alignment and enhance structural fidelity in complex regions.

\subsubsection{Ablation on Optimization Strategy}
We evaluate our dynamic optimization approach by comparing three training paradigms across two backbone architectures using the generated DIP dataset: supervised fine-tuning (SFT), standard DPO, and our FocusDPO methodology.
As shown in Tab.~\ref{tab:1}, FocusDPO consistently outperforms both SFT and conventional DPO across all metrics. For single-subject scenarios, FocusDPO achieves 0.751 under IPAP (8.7\% improvement over SFT, 8.1\% over DPO) and 0.802 under UNO (5.4\% improvement over SFT, 5.0\% over DPO). For multi-subject scenarios under UNO, FocusDPO achieves 0.570 (DINO), 0.739 (CLIP-I), and 0.326 (CLIP-T), representing improvements of 5.2\%/4.6\% (DINO), 1.0\%/1.1\% (CLIP-I), and 1.6\%/1.2\% (CLIP-T) over SFT/DPO respectively. These results demonstrate that improvements stem from our optimization strategy rather than dataset biases.

\section{Conclusion}
We present FocusDPO, a framework for multi-subject personalized image generation that tackles the challenge of preserving subject identity while avoiding attribute confusion. Our method introduces dynamic focus modulation mechanism that dynamically prioritize regions based on supervision complexity and semantic alignment.
Extensive experiments show that FocusDPO consistently improves pre-trained models, achieving state-of-the-art results on both single- and multi-subject benchmarks, setting a new standard for controllable multi-subject generation.

%% file: sections/appendix.tex
\section{Supplementary Ablation Study}
\subsection{Optimization Strategy Performance Analysis}
To evaluate the effectiveness of our FocusDPO, we conduct ablation studies across two architectural backbones: U-Net and Diffusion Transformer (DiT). For the U-Net architecture, we employ SDXL~\cite{sdxl} with IP-Adapter~\cite{ipadapter} as our baseline implementation.
\par
\begin{figure*}[htbp]
\centering
\includegraphics[width=1.0\linewidth]{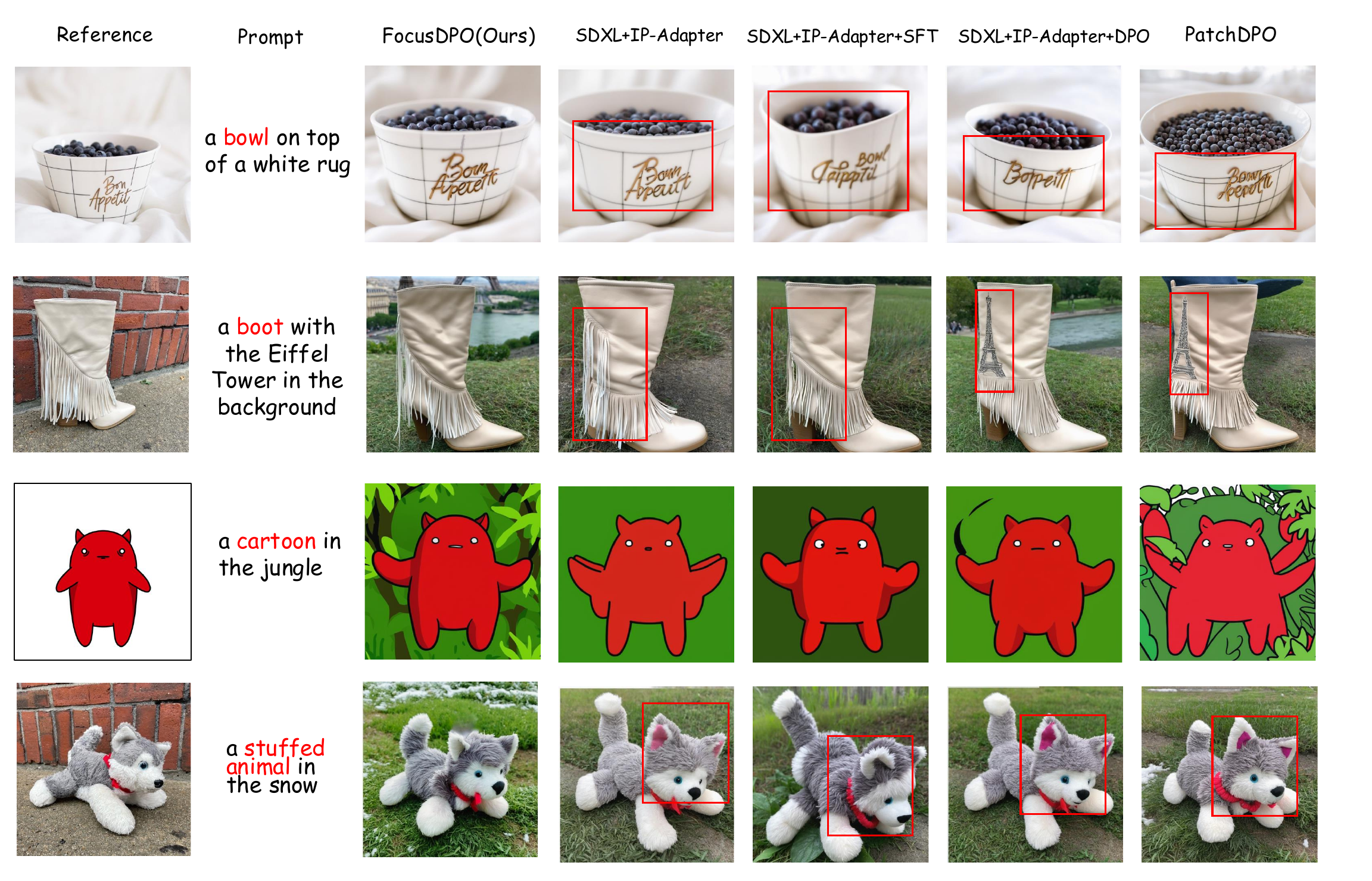} 
\caption{\textbf{Qualitative comparison of identity preservation across different optimization methods.}}
\label{fig:dpo}
\end{figure*}

\noindent
\textbf{Comparative Evaluation Framework.} Our ablation study isolates the contributions of the proposed optimization strategy from potential confounding factors including dataset artifacts and inherent DPO characteristics. The quantitative results presented in Section 4.3 are complemented by qualitative comparisons illustrated in Fig.~\ref{fig:dpo}, providing a comprehensive assessment across multiple evaluation dimensions.
As shown in Fig.~\ref{fig:dpo}, FocusDPO exhibits superior identity consistency preservation compared to existing optimization paradigms. The qualitative results reveal marked improvements across two critical dimensions:

\begin{itemize}
    \item \textbf{Semantic Fidelity}: Enhanced preservation of subject-specific characteristics and attributes
    \item \textbf{Visual Consistency}: Improved coherence in appearance and structural details across generated samples
\end{itemize}
\par
\noindent
\textbf{Comparison with Prior Work.} 
FocusDPO maintains robust identity preservation even when deployed on constrained architectures such as SDXL+IP-Adapter, indicating the generalizability of our optimization approach. Notably, FocusDPO substantially outperforms PatchDPO~\cite{patchdpo}—the sole prior work applying DPO principles to consistency-aware generation tasks. This performance advantage stems from our novel integration of:
\begin{itemize}
    \item Dynamic semantic attention mechanisms that adapt to content complexity
    \item Adaptive detail preservation strategies that maintain fine-grained visual fidelity
\end{itemize}
These findings validate both the theoretical foundations and practical efficacy of our approach across diverse architectural configurations.

\subsection{Analysis on $\mathbf{M}_{\text{prior}}$}
Building upon the ablation studies presented in Section 4.3, which examined the individual contributions of $\mathbf{M}_s$ and $\mathbf{M}_d$, we extend our analysis to investigate the influence of the prior mask $\mathbf{M}_{\text{prior}}$ on the overall framework performance. Our investigation reveals that $\mathbf{M}_{\text{prior}}$ exhibits a dual functionality: it serves as a superset constraint for Structure-Preserving Attention Field $\mathbf{M}_s$ while simultaneously constraining the Detail-Preserving Complexity Estimator $\mathbf{M}_d$. To systematically isolate and quantify the impact of $\mathbf{M}_{\text{prior}}$, we design three targeted ablation configurations.

\begin{figure*}[htbp]
\centering
\includegraphics[width=1.0\linewidth]{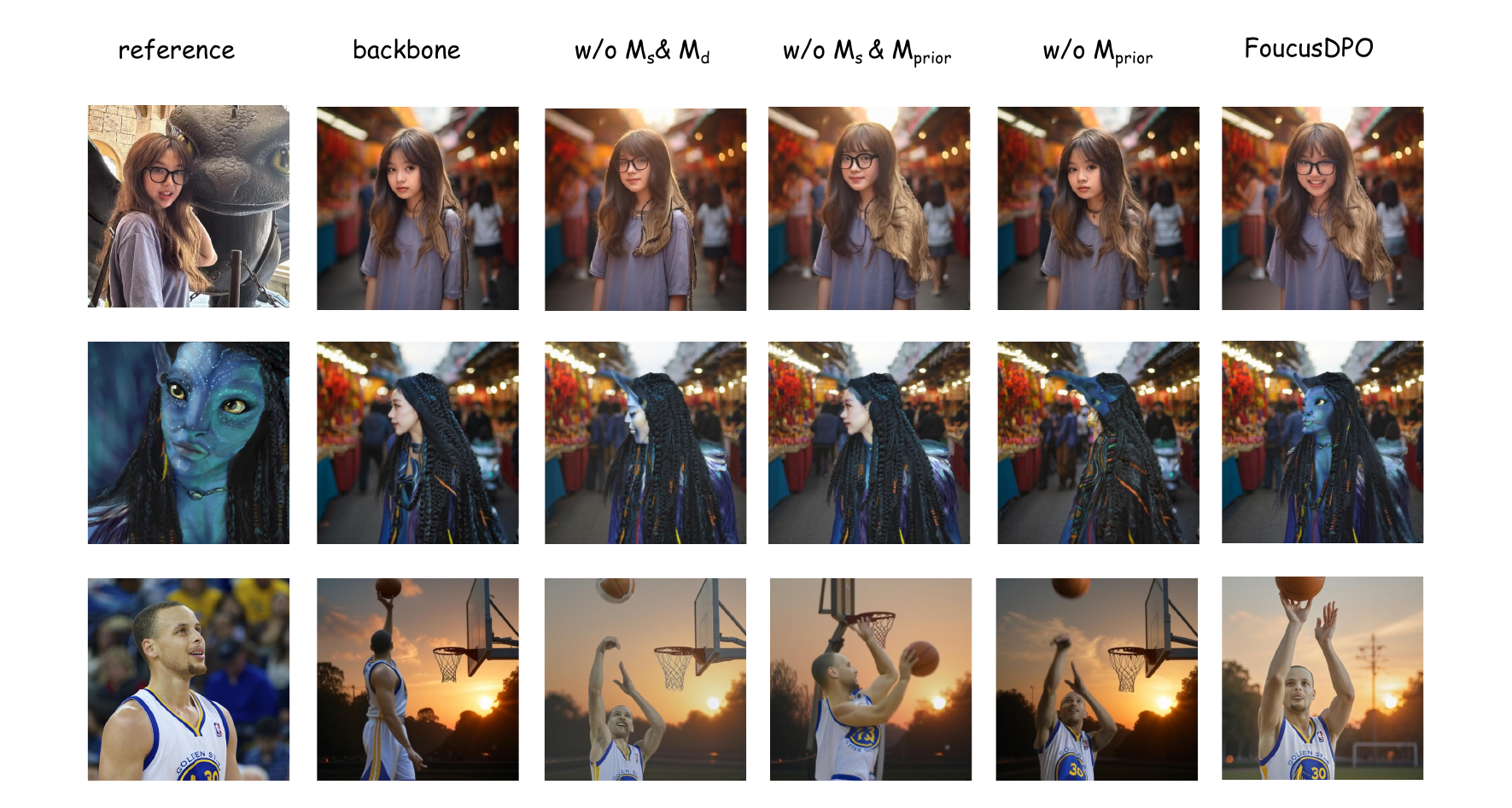} 
\caption{\textbf{Ablation study on the prior mask $\mathbf{M}_{\text{prior}}$ component.} The proposed FocusDPO outperforms all ablation variants, demonstrating the effectiveness of the integrated dynamic fusion optimization strategy.}
\label{fig:abla2}
\end{figure*}

\noindent\textbf{Configuration I: Prior-only baseline.} 
($w/o~\mathbf{M}_s \& \mathbf{M}_d
$) This configuration eliminates both semantic alignment and detail preservation components, focusing exclusively on the prior mask influence. The masking strategy reduces to $\mathbf{M} = \mathbf{M}_{\text{prior}}$, effectively providing a baseline for understanding the fundamental contribution of spatial priors.

\noindent\textbf{Configuration II: Information complexity Only.}
\noindent($w/o~\mathbf{M}_s \& \mathbf{M}_{\text{prior}}
$) This variant removes both the semantic structure alignment mask $\mathbf{M}_s$ and the spatial constraints imposed by $\mathbf{M}_{\text{prior}}$ on the information density mask, resulting in $\mathbf{M} = \mathbf{M}_d$. This configuration evaluates the effectiveness of information density guidance in isolation.

\noindent\textbf{Configuration III: Region Unconstrained information complexity.}~($w/o~\mathbf{M}_{\text{prior}}
$) This configuration retains the semantic mask $\mathbf{M}_s$ while removing the spatial constraints of $\mathbf{M}_{\text{prior}}$ on $\mathbf{M}_d$. The dynamic fusion strategy becomes:
\begin{equation}
\mathbf{M} = \begin{cases}
\mathbf{M}_s, & A_{\text{focus}} > \tau \\
\gamma \mathbf{M}_s + (1 - \gamma)\mathbf{M}_d , & A_{\text{focus}} \leq \tau
\end{cases}
\label{eq:prior_free}
\end{equation}

The comparative results presented in Fig.~\ref{fig:abla2} demonstrate that while each ablation configuration yields marginal improvements in consistency metrics, their performance remains substantially inferior to our complete FocusDPO formulation. These findings empirically validate the necessity of the integrated masking strategy and confirm the theoretical soundness of our proposed framework.

\subsection{Hyperparameter Sensitivity Analysis}

We conduct a systematic ablation study to analyze FocusDPO's sensitivity to two critical hyperparameters that govern the optimization dynamics:
\begin{itemize}
    \item \textbf{Focus Coverage Threshold ($\tau$):} Controls the transition mechanism from purely semantic-guided optimization to the integrated semantic-information density weighting scheme. This parameter determines when adaptive weighting mechanisms are activated during training.
    \item \textbf{Fusion Coefficient ($\gamma$):} Modulates the relative contributions between the semantic mask $\mathbf{M}_{\text{s}}$ and the information density mask $\mathbf{M}_{\text{d}}$ in the combined objective function, balancing semantic consistency and detail preservation.
\end{itemize}

\begin{figure*}[htbp]
\centering
\includegraphics[width=1.0\linewidth]{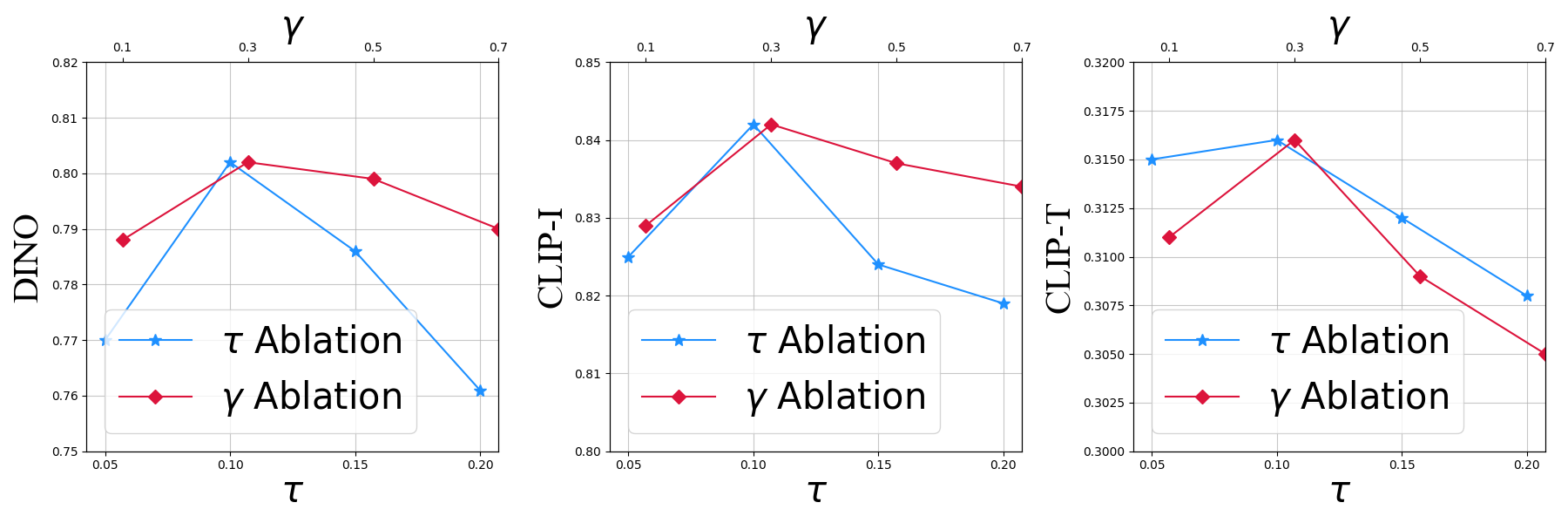} 
\caption{\textbf{Ablation study on the focus coverage threshold \(\tau\) and the fusion coefficient \(\gamma\).}}
\label{fig:hyperparam_ablation}
\end{figure*}

Our hyperparameter sweep evaluates performance across $\tau$ and $\gamma$ values, with validation performance measured using our standard evaluation metrics. The results are visualized in Fig.~\ref{fig:hyperparam_ablation}.
The empirical analysis reveals optimal convergence at:
\begin{itemize}
    \item $\tau = 0.1$: Facilitates activation of adaptive weighting at strategically appropriate training phases
    \item $\gamma = 0.3$: Establishes optimal equilibrium between semantic consistency constraints and information density regularization
\end{itemize}
 Our ablation reveals that FocusDPO exhibits stable performance within reasonable ranges of both hyperparameters, demonstrating the robustness of our approach. 

\section{DIP Dataset Construction and Quality Assurance}

\subsection{Dataset Overview and Structure}
The DIP (Disrupted-Instance Pair Dataset) dataset comprises 10,000 meticulously curated image triplets $(x_r, x_0^w, x_0^l)$, where $x_r$ denotes the reference image, $x_0^w$ represents the positive (winning) sample, and $x_0^l$ represents the negative (losing) sample. The dataset is systematically organized into two primary configurations:
\begin{itemize}
    \item \textbf{Single-subject subset}: 5,000 triplets focusing on individual subject preservation
    \item \textbf{Multi-subject subset}: 5,000 triplets addressing complex multi-subject scenarios
\end{itemize}
\noindent Each subset maintains balanced representation across three semantic categories: objects, animals, and humans (including real-person and anime styles), ensuring comprehensive coverage of diverse visual domains commonly encountered in personalized generation tasks.
\begin{figure*}[htbp]
\centering
\includegraphics[width=1.0\linewidth]{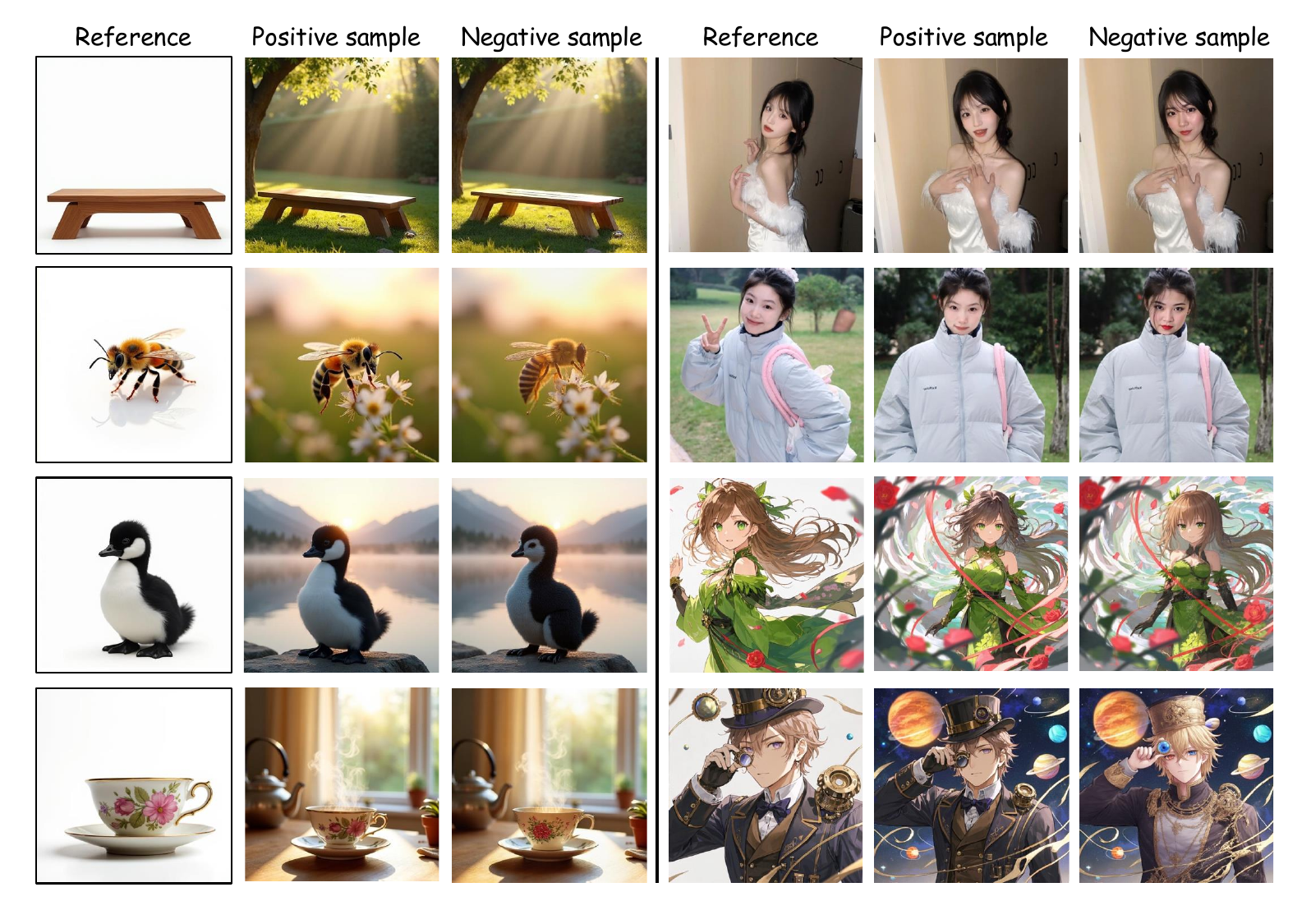} 
\caption{\textbf{Single-subject examples} from the DIP dataset across three content types: object, animal, human.}
\label{fig:single}
\end{figure*}

\subsection{Dataset Construction Methodology}
Our dataset construction follows a principled two-stage methodology:
\par 
\noindent 
\textbf{Stage 1: Reference-Target Correspondence Establishment}
We establish semantic correspondences between reference images and positive exemplars, ensuring high-quality alignment across key visual attributes including appearance and contextual elements.
\par 
\noindent
\textbf{Stage 2: Negative Sample Generation via Instance Disruption}
We apply systematic perturbation techniques to positive samples, deliberately diminishing subject-reference consistency while maintaining similar compositional structure and prompt descriptions. This approach creates challenging negative examples that share textual similarity with positive samples but exhibit poor visual alignment with references.

A critical insight from our dataset analysis reveals that positive and negative samples often correspond to nearly identical textual prompts, yet exhibit dramatically different levels of reference alignment. \textbf{Positive samples demonstrate strong semantic consistency} with preserved key features and subject positioning, while \textbf{negative samples show significant deviation} from reference characteristics despite similar descriptive content. This property creates region-aligned positive-negative pairs that provide robust training signals for our FocusDPO framework.

\begin{figure*}[htbp]
\centering
\includegraphics[width=0.9\linewidth]{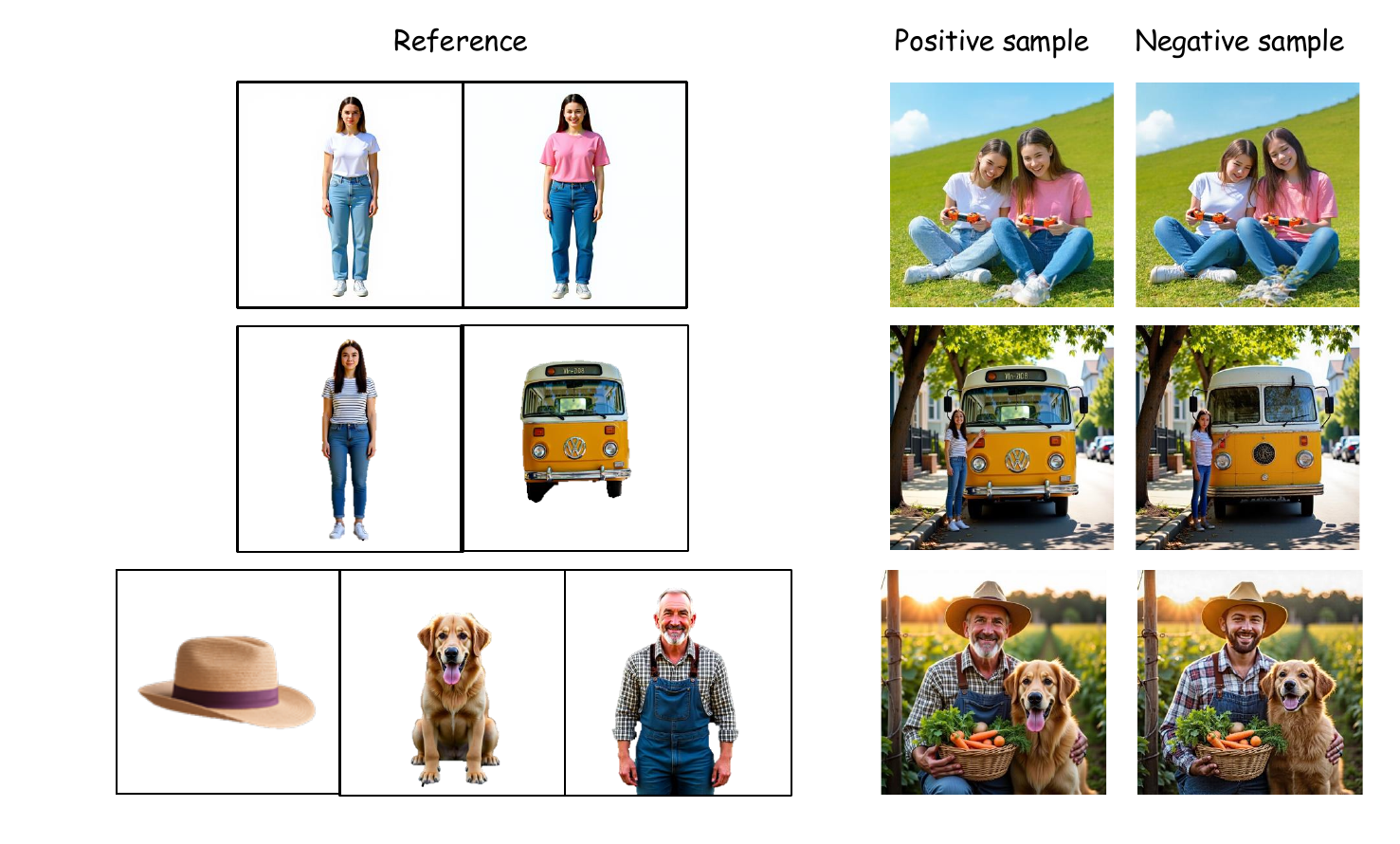} 
\caption{\textbf{Multi-subject examples} from the DIP dataset.}
\label{fig:multi}
\end{figure*}

\subsection{Quality Assurance Framework}
To ensure dataset reliability, we implement a comprehensive dual-validation framework combining automated assessment with expert human annotation:
\par 
\noindent
\textbf{Automated Evaluation Component:}
(1) Utilizes GPT-4o as the evaluation engine.
(2) Employs ten distinct evaluation queries per semantic category (as shown in Fig.~\ref{fig:standard}).
(3) Assesses subject-reference consistency across multiple semantic dimensions.
(4) Provides binary responses with scoring based on affirmative classifications.
\par 
\noindent
\textbf{Manual Annotation Component:}
(1) Expert annotators with specialized domain knowledge.
(2) Standardized evaluation guidelines replicating automated query structure.
(3) Consistent 10-point scoring scale across all evaluations.
(4) Independent assessment to eliminate potential automated bias.
\subsubsection{Consensus-Based Sample Selection}
Our sample inclusion criteria require \textbf{unanimous agreement} between both evaluation methodologies:
\begin{itemize}
    \item \textbf{Positive samples}: Both GPT-4o and human annotators must assign scores $\geq$ 9.
    \item \textbf{Negative samples}: Both evaluation approaches must yield scores $\leq$ 6.
    \item \textbf{Ambiguous cases}: Samples with divergent assessments are excluded to ensure dataset purity.
\end{itemize}

This stringent consensus requirement eliminates borderline cases and ensures that the final DIP dataset comprises only samples with robust consistency classifications, significantly enhancing the reliability of our training framework.

\subsection{Dataset Visualization and Analysis}
\subsubsection{Single-Subject Examples}
Fig.~\ref{fig:single} presents representative samples from our single-subject configuration, showcasing the diversity across object, animal, and human categories. These examples demonstrate the clear distinction between positive samples (maintaining strong reference alignment) and negative samples (exhibiting semantic drift while preserving similar compositional structure).
\subsubsection{Multi-Subject Examples}  
Fig.~\ref{fig:multi} exhibits multi-subject instantiations, highlighting the increased complexity of maintaining consistency across multiple reference subjects. These samples illustrate the dataset's capability to support training for challenging multi-subject scenarios where existing methods typically fail.

\subsubsection{Evaluation Standards}
\begin{figure*}[htbp]
\centering
\includegraphics[width=0.95\linewidth]{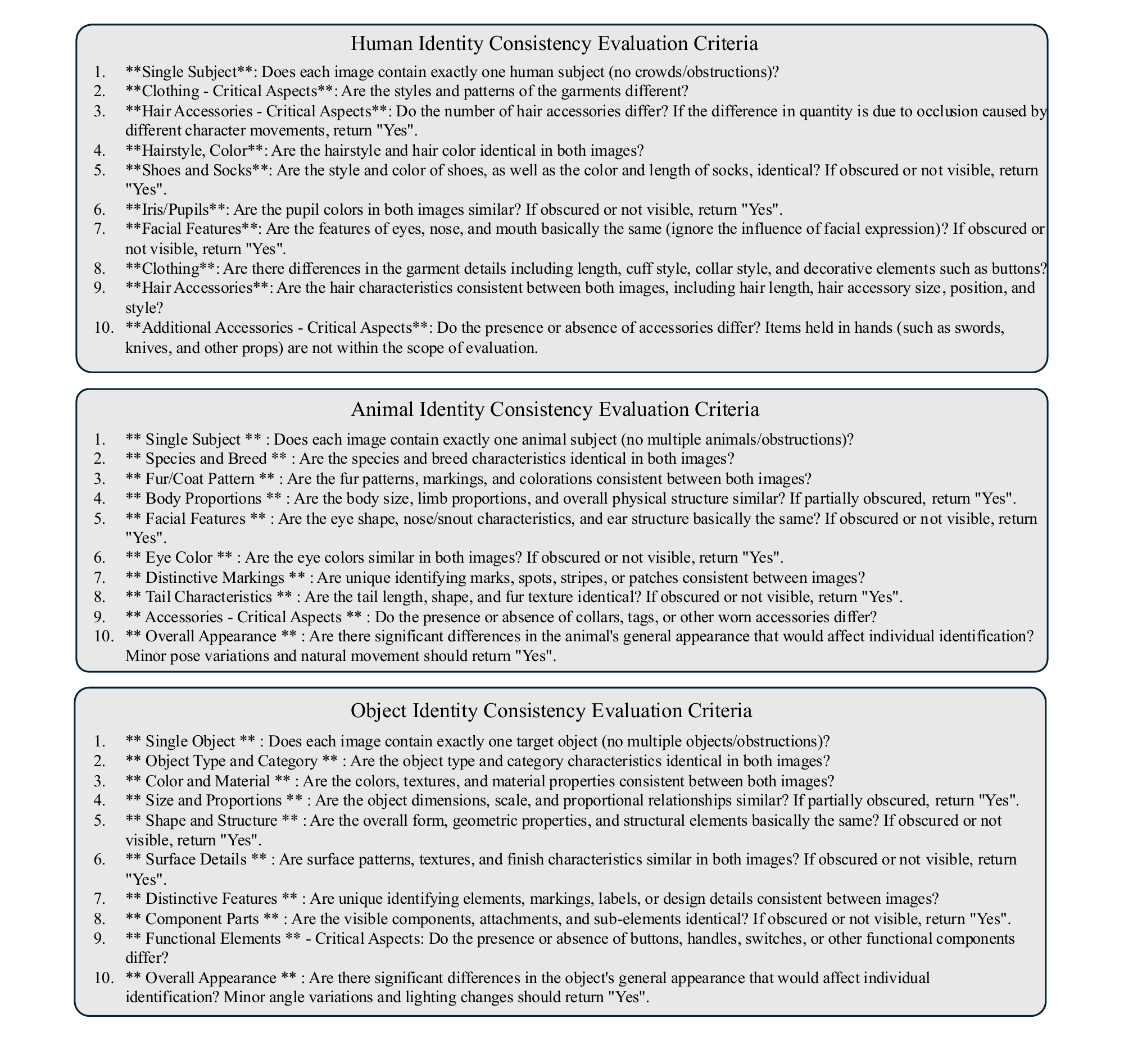} 
\caption{\textbf{Identity Consistency Evaluation Criteria of human, animal and object.}}
\label{fig:standard}
\end{figure*}

Fig.~\ref{fig:standard} details our identity consistency evaluation criteria across human, animal, and object categories, providing transparency in our quality assessment methodology and enabling reproducible dataset construction procedures.

